# Online Multi-view Clustering with Incomplete Views

Weixiang Shao*, Lifang He†, Chun-ta Lu* and Philip S. Yu*
*University of Illinois at Chicago, Chicago, IL
Email: {wshao4, clu29, psyu}@uic.edu
†Shenzhen University, Guangdong, China
Email: lifanghescut@gmail.com

*Abstract*—In the era of big data, it is common to have data with multiple modalities or coming from multiple sources, known as "multi-view data". Multi-view clustering provides a natural way to generate clusters from such data. Since different views share some consistency and complementary information, previous works on multi-view clustering mainly focus on how to combine various numbers of views to improve clustering performance. However, in reality, each view may be incomplete, i.e., instances missing in the view. Furthermore, the size of data could be extremely huge. It is unrealistic to apply multi-view clustering in large real-world applications without considering the incompleteness of views and the memory requirement. None of previous works have addressed all these challenges simultaneously. In this paper, we propose an online multi-view clustering algorithm, OMVC, which deals with large-scale incomplete views. We model the multi-view clustering problem as a joint weighted nonnegative matrix factorization problem and process the multi-view data chunk by chunk to reduce the memory requirement. OMVC learns the latent feature matrices for all the views and pushes them towards a consensus. We further increase the robustness of the learned latent feature matrices in OMVC via lasso regularization. To minimize the influence of incompleteness, dynamic weight setting is introduced to give lower weights to the incoming missing instances in different views. More importantly, to reduce the computational time, we incorporate a faster projected gradient descent by utilizing the Hessian matrices in OMVC. Extensive experiments conducted on four real data demonstrate the effectiveness of the proposed OMVC method.

*Keywords*-Multi-view clustering; Online algorithm; Incomplete views; Nonnegative matrix factorization

## I. INTRODUCTION

With the advance of technology, real data are often with multiple modalities or coming from multiple sources. Such data is called multi-view data. For example, in web image retrieval, the visual information of images and their textual tags can be regarded as two views; in web page clustering, a web page may be translated into multiple languages, each language can be seen as a view. Usually, multiple views provide consistent and complementary information for the semantically same data. By exploiting these characteristics between multi-view data, multi-view learning can obtain better performance of learning tasks than relying on just one single view [28].

Multi-view clustering [4], as one of basic tasks of multi-view learning, provides a natural way for generating clusters from multi-view data. A number of approaches have been proposed for multi-view clustering. Existing multi-view clustering algorithms can be roughly categorized into four categories [28]. Methods in the first category are based on subspace [8, 15, 20, 25], which learn a latent space so that different views are comparable in that space. Methods in the second category are co-training based algorithms [4, 14], which obtain the clustering results in an iterative manner. The third category aims to learn a unified similarity matrix among multi-view data, which serves as affinity matrix for final clustering [7, 23]. The last category is called late fusion [5, 9, 21]. Methods in this category first cluster each view independently and then combine the individual clusterings to produce a final clustering result.

Most of the above studies are based on the assumption that all of the views are complete, *i.e.,* each instance appears in all the views. However, due to the nature of the data or the cost of data collection, some views may suffer from the incompleteness of data (*i.e.,* instances within some views missing). In order to deal with this problem, different approaches have been explored [17, 25, 26, 29]. [29] is the first to deal with incomplete views by utilizing information from one complete view to refer to the kernel of incomplete views. [17, 26] are among the first attempts to solve multi-view clustering with none of the views complete. [25, 17] are the first attempts to solve multiple incomplete views clustering based on nonnegative matrix factorization (NMF). All the previous works require that the multi-view data can be fitted into the memory.

However, in reality, the size of data in multi-view may be extremely huge. For example, in Web scale data mining, one may encounter billions of Web pages and the dimension of the features may be as large as $\mathcal{O}(10^6)$. Data with such size and dimension clearly will not fit into the memory of a single machine. In fact, even a small corpus like Wikipedia has more than $3 \times 10^7$ pages in multiple languages. None of the existing multi-view clustering algorithms can handle data in such scale.

There are several challenges preventing us from applying multi-view clustering algorithms to large-scale data.
1) With the memory limitation, how to combine various views in different feature spaces and explore the consistency and complementary properties of different views to get better clustering solutions.
2) When the data are too large to fit into memory, how to deal with incomplete views, *i.e.*, how to minimize the influence of incompleteness of views.
3) How to effectively and efficiently learn the clustering solution even if the multi-view data are extremely large.

In this paper, we propose OMVC (**O**nline **M**ulti-**V**iew **C**lustering) to solve the above three challenges. To the best of our knowledge, this is the first online approach to solve the large-scale multi-view clustering problem with incomplete views. Basically, OMVC aims to reduce memory requirements by solving the problem in an online fashion. Instead of holding all the data in memory, it processes the multi-view data chunk by chunk. OMVC models the problem of multi-view clustering with incomplete views as a joint weighted nonnegative matrix factorization problem. It learns the latent feature matrices for incoming data chunk in each view and pushes them towards a common consensus. By storing the information of previous chunks in an aggregated form, OMVC does not store any specific previous data chunks. Inspired by the idea of weighted NMF [13], we use a dynamic weight setting to give lower weights to the incoming missing instances in different views. Thus, the proposed OMVC minimizes the negative influence from the missing instances. As one of the most commonly used regularization, $\ell_1$ (Lasso) regularization has been successfully applied in many algorithms [12, 17, 18]. By integrating weighted joint nonnegative matrix factorization and $\ell_1$ norm, OMVC tries to learn a latent subspace where the features of the same instance from different views will be co-regularized to a common consensus, while increasing the robustness of the learned latent feature matrices. More importantly, to reduce the computational time, OMVC incorporates a faster projected gradient descent algorithm by utilizing the Hessian matrices.

The contributions of this paper can be summarized as follows:
1) The proposed OMVC method is the first attempt to solve the problem of large-scale multi-view clustering with incomplete views in an online fashion.
2) We model the multi-view clustering as a joint nonnegative matrix factorization problem. The proposed method will capture the relation between different heterogeneous views and learn a consensus latent feature matrix across all the views.
3) We introduce lower weights for missing data in different views to reduce the influence of incomplete views. By using a dynamic weight setting, we can fill the incoming missing data with a quick estimation and give lower weights to the less informative estimations.
4) By utilizing lasso regularization, OMVC enforces the sparsity of latent feature matrices, and increase its robustness.
5) By doing multi-view clustering in an online fashion and adopting a faster projected gradient descent technique, OMVC can scale up to large data without appreciable sacrifice of performance.

The rest of this paper is organized as follows. In Section II, problem description and backgrounds are given. The details of the OMVC method are presented in Sections III and IV. Extensive experimental results and analysis are shown in Section V. Related works are discussed in Section VI and followed by conclusion in Section VII.

Table I: Summary of the Notations

| Notation | Description |
|---|---|
| $N$ | Total number of instances. |
| $n_v$ | Total number of views. |
| $D_v$ | Dimensionality of features in the $v$-th view. |
| $\mathbf{M}$ | Instance-view indicator matrix. |
| $\mathbf{X}^{(v)}$ | Data matrix for the $v$-th view. |
| $\mathbf{W}^{(v)}$ | Diagonal instance weight matrix for the $v$-th view. |
| $\mathbf{U}^{(v)}$ | Basis matrix for the $v$-th view. |
| $\mathbf{V}^{(v)}$ | Latent feature matrix for the data chunk in the $v$-th view. |
| $\mathbf{V}^*$ | Common consensus, latent feature matrix across all the views. |

## II. PRELIMINARIES

In this section, we will briefly describe the problem of online multi-view clustering with incomplete views. Then some background knowledge about clustering using nonnegative matrix factorization will be introduced. Table I summarizes the notations used throughout the paper.

### A. Problem Description

Assume that we are given $N$ instances in $n_v$ incomplete views $\{\mathbf{X}^{(v)}, v = 1, 2, ..., n_v\}$, where $\mathbf{X}^{(v)} \in \mathbb{R}^{D_v \times N}$ represents the data in the $v$-th view and $D_v$ is the dimensionality of features in the $v$-th view. Each view may be incomplete, *i.e.*, each of the view may have instances missing. We define an instance-view indicator matrix $\mathbf{M} \in \mathbb{R}^{N \times n_v}$ by

$$m_{i,j} = \begin{cases} 1 & \text{if the } i\text{-th instance is in the } j\text{-th view.} \\ 0 & \text{otherwise.} \end{cases} \quad (1)$$

where each column of $\mathbf{M}$ represents the instance presence in one view. Thus, in the incomplete views scenario, $\sum_{i=1}^{N} m_{i,j} < N$ for $j = 1, 2, ..., n_v$. Our goal is to partition all the $N$ instances into $K$ clusters by integrating all the $n_v$ incomplete views in an online fashion.

### B. Backgrounds

Let $\mathbf{X} \in \mathbb{R}_+^{D \times N}$ denote the nonnegative data matrix where each column represents an instance and each row represents a feature. Nonnegative matrix factorization (NMF) aims to factorize the data matrix $\mathbf{X}$ into two nonnegative matrices. We denote the two nonnegative matrices as $\mathbf{U} \in \mathbb{R}_+^{D \times R}$ and $\mathbf{V} \in \mathbb{R}_+^{N \times R}$. Here $R$ is the desired reduced dimensionality. To facilitate discussions, we call $\mathbf{U}$ the *basis matrix* and $\mathbf{V}$

the *latent feature matrix*. The objective function of NMF can be formulated as below:

$$\min_{\mathbf{U},\mathbf{V}} \mathcal{L} = \|\mathbf{X} - \mathbf{U}\mathbf{V}^T\|_F^2 \quad s.t. \ \mathbf{U} \geq 0, \mathbf{V} \geq 0, \quad (2)$$

where $\|\cdot\|_F$ is the Frobenius norm of the matrix. In clustering problems, the latent feature matrix $\mathbf{V}$ is used to extract the clustering solution. One option is to apply standard $K$-means on $\mathbf{V}$ to get the clustering solution. Another option is to add constraints to further restrict the rows of $\mathbf{V}$ and get the clustering indicators directly from $\mathbf{V}$. For example, we can constrain $\sum_j v_{i,j} = 1$ for every row $i$. Thus, $v_{i,j}$ will become the probability that the $i$-th instance belongs to cluster $j$.

In general, it is difficult to solve Eq. (2) as the objective function is not convex with $\mathbf{U}$ and $\mathbf{V}$ jointly. A common solution is to use an alternating way to update $\mathbf{U}$ and $\mathbf{V}$ [2]. One of the most well-known algorithms for implementing the alternating update rules is the multiplicative update approach in [16], which iteratively updates $\mathbf{U}$ and $\mathbf{V}$ by

$$\mathbf{U}_{i,j} \leftarrow \mathbf{U}_{i,j} \frac{(\mathbf{X}\mathbf{V})_{i,j}}{(\mathbf{U}\mathbf{V}^T\mathbf{V})_{i,j}} > 0 \quad \mathbf{V}_{i,j} \leftarrow \mathbf{V}_{i,j} \frac{(\mathbf{X}^T\mathbf{U})_{i,j}}{(\mathbf{V}\mathbf{U}^T\mathbf{U})_{i,j}} > 0$$

Another algorithm that solves this problem is *Projected Gradient Descent* (PGD) [19]. By fixing $\mathbf{V}$, PGD updates $\mathbf{U}$ using the first-order gradient:

$$\mathbf{U} \leftarrow P\left[\mathbf{U} - \gamma_k \nabla_\mathbf{U} \mathcal{L}(\mathbf{U}, \mathbf{V})\right] \quad (3)$$

where $\nabla_\mathbf{U}\mathcal{L}(\mathbf{U},\mathbf{V})$ is the gradient of $\mathcal{L}$ in Eq. (2) with respect to $\mathbf{U}$, $k$ is the index of the projected gradient iterations, $\gamma_k$ is the step size and $P$ is defined as

$$P[u_{i,j}] = \begin{cases} u_{i,j}, & \text{if } u_{i,j} \geq 0 \\ 0, & \text{otherwise.} \end{cases}$$

Similarly, PGD can be applied to update $\mathbf{V}$ with $\mathbf{U}$ fixed. PGD iteratively updates $\mathbf{U}$ and $\mathbf{V}$ until convergence. Both multiplicative update and PGD is proved to converge. However, both of them only use the first-order information, and thus the convergence rate is slow. To further accelerate the solving process, we borrow the idea from Newton's method [3] by utilizing the second-order information (*i.e.*, Hessian matrix) in our paper. Thus, the update equation for $\mathbf{U}$ in Eq. (3) becomes:

$$\mathbf{U} \leftarrow P\left[\mathbf{U} - \gamma_k \mathcal{H}^{-1}\left[\mathcal{L}(\mathbf{U},\mathbf{V})\right] \nabla_\mathbf{U}\mathcal{L}(\mathbf{U},\mathbf{V})\right], \quad (4)$$

where $\mathcal{H}^{-1}\left[\mathcal{L}(\mathbf{U},\mathbf{V})\right]$ is the inverse of the Hessian matrix. Similarly, we can apply the second-order PGD to update $\mathbf{V}$ with $\mathbf{U}$ fixed. We iteratively updates $\mathbf{U}$ and $\mathbf{V}$ until convergence.

III. ONLINE MULTI-VIEW CLUSTERING

The proposed online multi-view clustering algorithm processes the data in a streaming fashion with low computational and storage complexity. We will first describe how to derive the objective function.

*A. Objective of OMVC*

Given a set of incomplete multi-view data $\{\mathbf{X}^{(v)} \in \mathbb{R}_+^{D_v \times N}, v = 1, 2, ..., n_v\}$, we aim to find the latent feature matrices for each of the view and a common consensus, which represents the integrated information of all the views. The objective function can be written as below:

$$\min_{\{\mathbf{U}^{(v)}\},\{\mathbf{V}^{(v)}\},\mathbf{V}^*} \mathcal{L} = \sum_{v=1}^{n_v} \|\mathbf{X}^{(v)} - \mathbf{U}^{(v)}\mathbf{V}^{(v)T}\|_F^2 + \sum_{v=1}^{n_v} \alpha_v \|\mathbf{V}^{(v)} - \mathbf{V}^*\|_F^2$$

$$s.t. \ \mathbf{V}^* \geq 0, \mathbf{U}^{(v)} \geq 0, \mathbf{V}^{(v)} \geq 0, v = 1, 2, .., n_v. \quad (5)$$

where $\mathbf{U}^{(v)} \in \mathbb{R}_+^{D_v \times K}$ and $\mathbf{V}^{(v)} \in \mathbb{R}_+^{N \times K}$ are the basis matrix and latent feature matrix for the $v$-th view, $\mathbf{V}^* \in \mathbb{R}_+^{N \times K}$ is the consensus latent feature matrix across all the views, $K$ is the number of clusters and $\alpha_v$ is the trade-off parameter between reconstruction error and disagreement between view $v$ and the consensus.

Due to the incompleteness of each view, we cannot directly optimize the above objective function. One simple solution is to fill the missing instances with average value of the features first, and then solve the above objective function. However, this approach depends on the quality of the filled instances. For small incomplete percentages, the quality of the information contained in the filled features may be sufficient, whereas when the number of missing instance increases, the quality is often poor or even misleading. Thus, simply filling the missing instances cannot solve the problem. To eliminate the influence of the incomplete data, we borrow the idea from weighted NMF. We introduce a diagonal weight matrix $\mathbf{W}^{(v)} \in \mathbb{R}^{N \times N}$, whose diagonal element $w_{i,i}^{(v)}$ represents the weight of the $i$-th instance in the $v$-th view. We give weight 1 to the instances that appear in the view, and give lower weight to the missing instances (average filled instances) in the view. We will discuss how to dynamically adjust the weight later in this section. The objective function after adding the weight matrices is:

$$\mathcal{L} = \sum_{v=1}^{n_v} \|(\mathbf{X}^{(v)} - \mathbf{U}^{(v)}\mathbf{V}^{(v)T})\mathbf{W}^{(v)}\|_F^2 + \sum_{v=1}^{n_v} \alpha_v \|\mathbf{W}^{(v)}(\mathbf{V}^{(v)} - \mathbf{V}^*)\|_F^2 \quad (6)$$

By assigning different weights to instances in difference views, we can give larger weights to more informative estimations of the missing instances and lower weights to less informative or misleading estimations. Additionally, considering the nature of incomplete views, we adopt $\ell_1$ norm to enforce the sparsity of the latent feature matrix, which is robust to noises and outliers and widely used in many algorithms [12, 17].

$$\mathcal{L} = \sum_{v=1}^{n_v} \|(\mathbf{X}^{(v)} - \mathbf{U}^{(v)}\mathbf{V}^{(v)T})\mathbf{W}^{(v)}\|_F^2$$
$$+ \sum_{v=1}^{n_v} \alpha_v \|\mathbf{W}^{(v)}(\mathbf{V}^{(v)} - \mathbf{V}^*)\|_F^2 + \sum_{v=1}^{n_v} \beta_v \|\mathbf{V}^{(v)}\|_1 \quad (7)$$

where $\|\cdot\|_1$ is the $\ell_1$ norm and $\beta_v$ is the trade-off parameter between the sparsity and accuracy of reconstruction for the $v$-th view.

In real-world applications, the data matrices may be too large to fit into the memory. We propose to solve the above optimization problem in an online/streaming fashion with low computational and storage complexity. Note that the objective function $\mathcal{L}$ can be decomposed as:

$$\mathcal{L} = \sum_{v=1}^{n_v}\sum_{i=1}^{N} \|w_{i,i}^{(v)}(\mathbf{x}_i^{(v)} - \mathbf{U}^{(v)}\mathbf{v}_i^{(v)})\|_F^2 \\ + \sum_{v=1}^{n_v}\sum_{i=1}^{N} \alpha_v \|w_{i,i}^{(v)}(\mathbf{v}_i^{(v)} - \mathbf{v}_i^*)\|_F^2 + \sum_{v=1}^{n_v}\sum_{i=1}^{N} \beta_v \|\mathbf{v}_i^{(v)}\|_1 \quad (8)$$

where $\mathbf{x}_i^{(v)}$ is the $i$-th column of $\mathbf{X}^{(v)}$ and $\mathbf{v}_i^{(v)} \in \mathbb{R}^K$ is the $i$-th column of $\mathbf{V}^{(v)T}$. Clearly, when all the basis matrices $\mathbf{U}^{(v)}$ are fixed, the calculation of $\mathbf{v}_i^{(v)}$ and $\mathbf{v}_i^*$ is independent for different $i$. This property would allow us approximate the optimal solution by processing the data one by one (chunk by chunk). Let's split the input data into chunks and at time $t$, we process a chunk of data points in all the views $\{\mathbf{X}_t^{(v)} \in \mathbb{R}^{s \times D_v}\}$, where $s$ is the size of the data chunk (number of instances). Eq. (8) can be written as:

$$\mathcal{L} = \sum_{v=1}^{n_v}\sum_{t=1}^{\lceil N/s \rceil} \|(\mathbf{X}_t^{(v)} - \mathbf{U}^{(v)}\mathbf{V}_t^{(v)T})\mathbf{W}_t^{(v)}\|_F^2 \\ + \sum_{v=1}^{n_v}\sum_{t=1}^{\lceil N/s \rceil} \alpha_v \|\mathbf{W}_t^{(v)}(\mathbf{V}_t^{(v)} - \mathbf{V}_t^*)\|_F^2 + \sum_{v=1}^{n_v}\sum_{t=1}^{\lceil N/s \rceil} \beta_v \|\mathbf{V}_t^{(v)}\|_1 \quad (9)$$

where $\mathbf{X}_t^{(v)}$ is the $t$-th data chunk in the $v$-th view, $\mathbf{V}_t^{(v)} \in \mathbb{R}^{s \times K}$ is the latent feature matrix for the $t$-th data chunk, and $\mathbf{W}_t^{(v)} \in \mathbb{R}^{s \times s}$ is the diagonal weight matrix for the $t$-th data chunk.

### B. Dynamic Weight Setting for Missing Instances

In the previous discussion, we mentioned that we will fill the missing instances with the average values of the features and assign different weights to them. We would like to assign lower weights to the less informative estimations (averaged instances) and higher weights to the more informative estimations. However, since the entire data cannot be held in memory, the data can only be read in a streaming fashion. Thus, the average values cannot be directly calculated. Instead of filling the missing instances with the global average values, we fill the missing instances with the dynamic (up-to-date) average when we read in a new data point/chunk:

$$\mathbf{x}_t^{(v)} = \frac{\sum_{i=1}^{t} m_{i,v} \mathbf{x}_i^{(v)}}{\sum_{i=1}^{t} m_{i,v}}, \quad (10)$$

which can be calculated efficiently for every incoming missing instances. The weight $w_{t,t}^{(v)}$ is set dynamically to the up-to-date percentage of the available instances in view $v$:

$$w_{t,t}^{(v)} = \begin{cases} 1 & \text{if instance } t \text{ appears in view } v. \\ \frac{\sum_{i=1}^{t} m_{i,v}}{t} & \text{otherwise.} \end{cases} \quad (11)$$

We can observe that $w_{t,t}^{(v)}$ is lower if the estimate of $\mathbf{x}_t^{(v)}$ is made under a higher percentage of missing instances. Thus, $w_{t,t}^{(v)}$ represents the quality of the estimated average features. Next, we will describe how to optimize the objective function in an online fashion.

## IV. OPTIMIZATION ALGORITHMS

In this section, we first solve the objective function of OMVC derived in the previous section. Then, we will discuss the one-pass OMVC and the multi-pass OMVC algorithms. The convergence and complexity analysis will be given in the end of this section.

### A. Solution

From Eq. (9) we can see that at each time $t$, we need to optimize $\{\mathbf{U}^{(v)}\}$, $\{\mathbf{V}_t^{(v)}\}$ and $\mathbf{V}_t^*$. However, the objective function is not jointly convex, so we have to update $\{\mathbf{U}^{(v)}\}$, $\{\mathbf{V}_t^{(v)}\}$ and $\mathbf{V}_t^*$ in an alternating way. Thus, there are three subproblems in OMVC described as follows.

*1) Optimize $\{\mathbf{U}^{(v)}\}$ with $\{\mathbf{V}_t^{(v)}\}$ and $\mathbf{V}_t^*$ Fixed:* From Eq. (9) we can observe that the optimization of $\mathbf{U}^{(v)}$ is independent for different $v$ with $\{\mathbf{V}_t^{(v)}\}$ and $\mathbf{V}_t^*$ fixed. To optimize $\mathbf{U}^{(v)}$ for a specific view $v$ at time $t$, we only need to minimize the following objective:

$$\mathcal{J}^{(t)}(\mathbf{U}^{(v)}) = \sum_{i=1}^{t} \|(\mathbf{X}_i^{(v)} - \mathbf{U}^{(v)}\mathbf{V}_i^{(v)T})\mathbf{W}_i^{(v)}\|_F^2 \\ \text{s.t. } \mathbf{U}^{(v)} \geq 0 \quad (12)$$

Taking the first-order derivative, the gradient of $\mathcal{J}^{(t)}$ with respect to $\mathbf{U}^{(v)}$ is

$$\nabla \mathcal{J}^{(t)}(\mathbf{U}^{(v)}) = 2\mathbf{U}^{(v)}\sum_{i=1}^{t} \mathbf{V}_i^{(v)T}\tilde{\mathbf{W}}_i^{(v)}\mathbf{V}_i^{(v)} - 2\sum_{i=1}^{t} \mathbf{X}_i^{(v)}\tilde{\mathbf{W}}_i^{(v)}\mathbf{V}_i^{(v)} \quad (13)$$

Here, $\tilde{\mathbf{W}}_i^{(v)} = \mathbf{W}_i^{(v)T}\mathbf{W}_i^{(v)} = \mathbf{W}_i^{(v)}\mathbf{W}_i^{(v)T}$. For the sake of convenience, we introduce two terms $\mathbf{A}_t^{(v)}$ and $\mathbf{B}_t^{(v)}$:

$$\mathbf{A}_t^{(v)} = \sum_{i=1}^{t} \mathbf{V}_i^{(v)T}\tilde{\mathbf{W}}_i^{(v)}\mathbf{V}_i^{(v)} \quad \mathbf{B}_t^{(v)} = \sum_{i=1}^{t} \mathbf{X}_i^{(v)}\tilde{\mathbf{W}}_i^{(v)}\mathbf{V}_i^{(v)}$$

which can be computed incrementally with low storage. Thus, Eq. (13) can be written as:

$$\nabla \mathcal{J}^{(t)}(\mathbf{U}^{(v)}) = 2\mathbf{U}^{(v)}\mathbf{A}_t^{(v)} - 2\mathbf{B}_t^{(v)} \quad (14)$$

Therefore, the Hessian matrix of $\mathcal{J}^{(t)}(\mathbf{U}^{(v)})$ with respect to $\mathbf{U}^{(v)}$ is

$$\mathcal{H}\left[\mathbf{U}^{(v)}\right] = 2\mathbf{A}_t^{(v)} \quad (15)$$

Using the second order PGD, the update equation for $\mathbf{U}^{(v)}$ at time $t$ is:

$$\mathbf{U}_{k+1}^{(v)} \leftarrow P\left[\mathbf{U}_k^{(v)} - \gamma_k \nabla \mathcal{J}^{(t)}(\mathbf{U}_k^{(v)})\mathcal{H}^{-1}[\mathbf{U}_k^{(v)}]\right] \quad (16)$$

where $k$ is the number of iterations, and $\gamma_k$ is the step size.

For choosing an appropriate step size $\gamma_k$, we consider the simple and effective Armijo rule along the projection described in [2], that is, $\gamma_k = \eta^{\varphi_k}$, and $\varphi_k$ is the first non-negative integer such that

$$\mathcal{J}^{(t)}(\mathbf{U}_{k+1}^{(v)}) - \mathcal{J}^{(t)}(\mathbf{U}_k^{(v)}) \leq \sigma \langle \nabla \mathcal{J}^{(t)}(\mathbf{U}_k^{(v)}), \mathbf{U}_{k+1}^{(v)} - \mathbf{U}_k^{(v)} \rangle \quad (17)$$

where $\sigma \in (0,1)$ and $\langle \cdot, \cdot \rangle$ is the sum of the component-wise product of two matrices. The condition (17) ensures that the sufficient decrease of the function value per iteration. Bertsekas [3] has proved that by selecting the step sizes $1, \beta, \beta^2, \cdots, \gamma_k > 0$ satisfying (17) always exists and every limit point of $\{\mathbf{U}_k^{(v)}\}$ is a stationary point of (12).

Following [19], to reduce the computational cost, inequality (17) can be reformulated as:

$$(1-\sigma)\langle \nabla \mathcal{J}^{(t)}(\mathbf{U}_k^{(v)}), \mathbf{U}_{k+1}^{(v)} - \mathbf{U}_k^{(v)} \rangle \\ + \frac{1}{2}\langle \mathbf{U}_{k+1}^{(v)} - \mathbf{U}_k^{(v)}, \mathcal{H}(\mathbf{U}_k^{(v)})(\mathbf{U}_{k+1}^{(v)} - \mathbf{U}_k^{(v)}) \rangle \leq 0 \quad (18)$$

*2) Optimize* $\{\mathbf{V}_t^{(v)}\}$ *with* $\mathbf{V}_t^*$ *and* $\{\mathbf{U}^{(v)}\}$ *Fixed:* Given $\mathbf{V}_t^*$ and $\{\mathbf{U}^{(v)}\}$ fixed, the optimization of $\{\mathbf{V}_t^{(v)}\}$ is independent for different $v$. To optimize $\mathbf{V}_t^{(v)}$ for the $v$-th view, we only need to minimize the following objective:

$$\mathcal{J}(\mathbf{V}_t^{(v)}) = \|(\mathbf{X}_t^{(v)} - \mathbf{U}^{(v)}\mathbf{V}_t^{(v)T})\mathbf{W}_t^{(v)}\|_F^2 \\ + \alpha_v \|\mathbf{W}_t^{(v)}(\mathbf{V}_t^{(v)} - \mathbf{V}_t^*)\|_F^2 + \beta_v \|\mathbf{V}_t^{(v)}\|_1 \quad (19) \\ \text{s.t. } \mathbf{V}_t^{(v)} \geq 0$$

Taking the first-order derivative, the gradient of $\mathcal{J}$ with respect to $\mathbf{V}_t^{(v)}$ is

$$\nabla \mathcal{J}(\mathbf{V}_t^{(v)}) = 2\tilde{\mathbf{W}}_t^{(v)}(\mathbf{V}_t^{(v)}\mathbf{U}^{(v)T} - \mathbf{X}^{(v)T})\mathbf{U}^{(v)} \\ + 2\alpha_v \tilde{\mathbf{W}}_t^{(v)}(\mathbf{V}_t^{(v)} - \mathbf{V}_t^*) + \beta_v \mathbf{1} \quad (20)$$

Let $\mathbf{v}_{ti}^{(v)}$ be the $i$-th column of $\mathbf{V}_t^{(v)T}$ and $w_{ti,ti}^{(v)}$ be the $i$-th diagonal element of $\mathbf{W}_t^{(v)}$, then the Hessian matrix of $\mathcal{J}(\mathbf{v}_{ti}^{(v)})$ with respect to $\mathbf{v}_{ti}^{(v)}$ is

$$\mathcal{H}\left[\mathbf{v}_{ti}^{(v)}\right] = 2{w_{ti,ti}^{(v)}}^2(\mathbf{U}^{(v)T}\mathbf{U}^{(v)} + \alpha_v I_K) \quad (21)$$

Using the second-order PGD, the update equation for $\mathbf{v}_{ti}^{(v)}$ is:

$$\mathbf{v}_{ti}^{(v)} \leftarrow P\left[\mathbf{v}_{ti}^{(v)} - \gamma \mathcal{H}^{-1}[\mathbf{v}_{ti}^{(v)}]\nabla \mathcal{J}(\mathbf{v}_{ti}^{(v)})\right] \quad (22)$$

Here, $\gamma$ is the step size. We use the similar search procedure as the previous subsection to find the step size that satisfies the Armijo rule.

*3) Optimize* $\mathbf{V}_t^*$ *with* $\{\mathbf{U}^{(v)}\}$ *and* $\{\mathbf{V}_t^{(v)}\}$ *Fixed:* To optimize $\mathbf{V}_t^*$ with $\{\mathbf{V}_t^{(v)}\}$ and $\{\mathbf{U}^{(v)}\}$ fixed, we only need to minimize the following objective function:

$$\mathcal{J}(\mathbf{V}_t^*) = \sum_{v=1}^{n_v} \alpha_v \|\mathbf{W}_t^{(v)}(\mathbf{V}_t^{(v)} - \mathbf{V}_t^*)\|_F^2 \quad \text{s.t. } \mathbf{V}_t^* \geq 0 \quad (23)$$

---

**Algorithm 1:** One-pass OMVC with mini-batch mode.

**Input**: Data matrices of all the incomplete views $\{\mathbf{X}^{(v)}\}$. The number of clusters $K$, the batch size $s$. Parameters $\{\alpha_v\}$ and $\{\beta_v\}$.

1    $\mathbf{A}_0^{(v)} = \mathbf{0}$, $\mathbf{B}_0^{(v)} = \mathbf{0}$ for each view $v$. **for** $t = 1 : \lceil N/s \rceil$ **do**
2      Draw $\mathbf{X}_t^{(v)}$ for all the views.
3      Fill in the missing instances and set the weights. **repeat**
4        **for** $v = 1 : n_v$ **do**
5          Update $\mathbf{U}^{(v)}$ according to Eq. (16).
6          Update $\mathbf{V}_t^{(v)}$ according to Eq. (22).
7        Calculate $\mathbf{V}_t^*$ according to Eq. (25).
8      **until** *Convergence*;
9      $\mathbf{A}_t^{(v)} = \mathbf{A}_{t-1}^{(v)} + \mathbf{V}_t^{(v)T}\tilde{\mathbf{W}}_t^{(v)}\mathbf{V}_t^{(v)}$
10     $\mathbf{B}_t^{(v)} = \mathbf{B}_{t-1}^{(v)} + \mathbf{X}_t^{(v)}\tilde{\mathbf{W}}_t^{(v)}\mathbf{V}_t^{(v)T}$
11     Extract clustering solution from $\mathbf{V}_t^*$

---

Here, we assume that $\alpha_v$ is positive. Taking the derivative of the objective $\mathcal{J}$ in Eq. (23) over $\mathbf{v}_t^*$ and set it to 0:

$$\frac{\partial \mathcal{J}}{\partial \mathbf{V}_t^*} = \sum_{v=1}^{n_v} 2\alpha_v \tilde{\mathbf{W}}_t^{(v)}\mathbf{V}_t^* - \sum_{v=1}^{n_v} 2\alpha_v \tilde{\mathbf{W}}_t^{(v)}\mathbf{V}_t^{(v)} = 0 \quad (24)$$

Since $\mathbf{W}_t^{(v)}$ is a positive diagonal matrix and $\alpha_v$ is positive, $\sum_{v=1}^{n_v} \alpha_v \tilde{\mathbf{W}}_t^{(v)} = \sum_{v=1}^{n_v} \alpha_v \mathbf{W}_t^{(v)T}\mathbf{W}_t^{(v)}$ is also a positive diagonal matrix, whose inverse can be quickly calculated. Solving Eq. (24), we have an exact solution for $\mathbf{V}_t^*$:

$$\mathbf{V}_t^* = \left(\sum_{v=1}^{n_v} \alpha_v \tilde{\mathbf{W}}_t^{(v)}\right)^{-1} \sum_{v=1}^{n_v} \alpha_v \tilde{\mathbf{W}}_t^{(v)}\mathbf{V}_t^{(v)} \geq 0 \quad (25)$$

### B. One-Pass OMVC

The complete one-pass algorithm procedure is shown in Algorithm 1. Several important points need to be noted. First, at each time $t$, we do not need to recompute new $\mathbf{A}_t^{(v)}$ and $\mathbf{B}_t^{(v)}$. We only need to compute $\mathbf{V}_t^{(v)T}\tilde{\mathbf{W}}_t^{(v)}\mathbf{V}_t^{(v)}$ and $\mathbf{X}_t^{(v)}\tilde{\mathbf{W}}_t^{(v)}\mathbf{V}_t^{(v)T}$, and add them to old $\mathbf{A}_{t-1}^{(v)}$ and $\mathbf{B}_{t-1}^{(v)}$.

$$\mathbf{A}_t^{(v)} = \mathbf{A}_{t-1}^{(v)} + \mathbf{V}_t^{(v)T}\tilde{\mathbf{W}}_t^{(v)}\mathbf{V}_t^{(v)} \quad (26)$$

$$\mathbf{B}_t^{(v)} = \mathbf{B}_{t-1}^{(v)} + \mathbf{X}_t^{(v)}\tilde{\mathbf{W}}_t^{(v)}\mathbf{V}_t^{(v)} \quad (27)$$

Second, in the algorithm, we need to calculate the inverse of Hessian matrix (with dimension $K \times K$) in the iteration. The computation cost of inverse of a matrix will be high if the number of clusters $K$ becomes large. However, in most of the cases, the number of clusters is limited to a small number. Even if $K$ is very large in some cases, we can use other ways to approximate the inverse, such as the diagonal approximation [30]. Third, the proposed alternative update procedures for $\{\mathbf{U}^{(v)}, \mathbf{V}_t^{(v)}, \mathbf{V}_t^*\}$ converge. The proof of convergence is shown in Section IV-D.

## C. Multi-Pass OMVC

In data stream scenario, often only one pass over the data is available. In many other applications, it is feasible to do multiple passes. In the one-pass OMVC, the consensus latent feature matrix $\mathbf{V}^*$, which represents the clustering assignment/possibility, is computed in a sequential greedy way. It is expected that for the data points that come first, the performance of clustering may not be satisfactory. However, multi-pass OMVC gives a chance to improve the performance for those earlier data points.

In the multi-pass OMVC, $\{\mathbf{V}^{(v)}\}$ and $\mathbf{V}^*$ can be updated using $\{\mathbf{U}^{(v)}\}$ in the previous pass. $\{\mathbf{A}^{(v)}\}$ and $\{\mathbf{B}^{(v)}\}$ from previous pass can be used and updated. Also, the weights for missing instances will be more accurate after the first pass. Thus, the performance of clustering after multiple passes is expected to be better than that of one-pass OMVC.

## D. Convergence and Complexity

The convergence of the proposed algorithm OMVC can be illustrated by the following theorem.

THEOREM *1:* Any limit point of the sequence $\{\mathbf{U}_k^{(v)}, \mathbf{V}_{tk}^{(v)}, \mathbf{V}_{tk}^*\}$ generated by Algorithm 1 is a stationary point of Eq. (19).

*Proof:* The Algorithm 1 can be viewed as the "block coordinate descent" method in bound-constrained optimization [2], where sequentially one block of variables is minimized under corresponding constraints and the remaining two blocks are fixed. Regarding the convergence of "block coordinate descent" methods, Grippo and Sciandrone [10] have shown that for the case of *three blocks*, the convergence can be ensured by requiring only the strict quasiconvexity of the objective function with respect to one component. Clearly, we can satisfy this condition over $\mathbf{V}^{*(t)}$. Therefore, the proof of Theorem 1 is an immediate consequence of Proposition 5 of [10]. ∎

Next, we discuss the computational complexity of OMVC algorithm. There are three subproblems for OMVC algorithm: optimizing $\mathbf{U}^{(v)}$, optimizing $\mathbf{V}_t^{(v)}$ and optimizing $\mathbf{V}_t^*$.

To optimize $\mathbf{U}^{(v)}$, we need to calculate the gradient $\nabla \mathcal{J}(\mathbf{U}^{(v)})$ and the Hessian matrix $\mathcal{H}[\mathbf{U}^{(v)}]$. Assume that $K \ll D_v$ and $K < s$. From Eq. (14) and Eq. (15), we can see that the complexity is $\mathcal{O}(KD_v s)$ for the gradient and $\mathcal{O}(K^2 s)$ for the Hessian, where $K$ is the number of clusters, $D_v$ is the feature dimension in the $v$-th view and $s$ is the size of data chunk. Also, the complexity to update $\mathbf{U}^{(v)}$ using gradient and Hessian is $\mathcal{O}(K^3 + D_v K^2)$, where $K^3$ is the time complexity for the inversion of Hessian matrix. Since $K$ is usually very small, the inverse will be done very quickly. So the complexity for updating $\mathbf{U}^{(v)}$ once is $\mathcal{O}(KD_v s)$. According to [19], the complexity for searching the step size satisfying Eq. (18) is $t_{in}K^2 D_v$, where $t_{in}$ is the number of iterations to find the step size. So the overall complexity for updating $\mathbf{U}^{(v)}$ is $\mathcal{O}(KD_v s + t_{in}K^2 D_v) < \mathcal{O}(t_{in}KD_v s)$.

Similarly, we can find that the overall time complexity for updating $\mathbf{V}_t^{(v)}$ is $\mathcal{O}(KD_v s + t_{in}K^2 s) < \mathcal{O}(t_{in}KD_v s)$.

Another subproblem of OMVC is optimizing $\mathbf{V}_t^*$. From Eq. (25), we can see that it takes $\mathcal{O}(n_v K s)$ to calculate $\mathbf{V}_t^*$, where $n_v$ is the number of views. Thus, the complexity for one data chunk in one pass will be $\mathcal{O}(t_{out}t_{in}n_v KDs + t_{out}n_v Ks) = \mathcal{O}(t_{out}t_{in}n_v KDs)$, where $t_{out}$ is the average number of iterations to update $\mathbf{V}_t^{(v)}$ and $\mathbf{U}^{(v)}$ and $D$ is the average feature dimension for multiple views. Thus, the overall time complexity of one pass OMVC is $\mathcal{O}(t_{out}t_{in}n_v KDN)$, where $N$ is the total number of instances.

The complexity of some most recent NMF based off-line multi-view clustering methods such as MultiNMF [20] and MIC [25] are $\mathcal{O}(t_{out}t_{in}n_v KDN)$, which is in the same order as OMVC. However, all the off-line algorithms require $\mathcal{O}(n_v DN)$ memory space. When the data are too big to fit in the memory, the off-line algorithms will not work. The proposed OMVC only requires $\mathcal{O}(n_v Ds)$ memory space ($s \ll N$), which makes it work for really large data.

When the data size is large, the IO cost can be a significant (sometimes dominating) portion of the total cost. Our experiment results will verify that we often do not need many passes to achieve very accurate results.

## V. EXPERIMENT

### A. Dataset

In this paper, two small datasets and two large datasets are used to evaluate the proposed method OMVC. The summary of the datasets is shown in Table II, and the details of the datasets are as follows:

- **Web Knowledge Base (WebKB)**[1]: It is a subset of web documents from four universities [27]. The data contains 1,051 documents from two classes, course or non-course. Each document has two views: 1) the textual content of the web page and 2) the anchor text on links in other web pages pointing to the web page.
- **Handwritten Dutch Digit Recognition (Digit)**[2]: This data contains 2,000 handwritten numerals ("0"-"9") extracted from a collection of maps. Five views are used in our experiments: (1) 76 Fourier coefficients of the character shapes, (2) 216 profile correlations, (3) 64 Karhunen-Loeve coefficients, (4) 240 pixel averages in $2 \times 3$ windows, (5) 47 Zernike moments.
- **Reuters Multilingual Text Data (Reuters)**[3]: This data contains features of 111,740 documents originally written in five different languages (English, French, German, Spanish and Italian), and their translations, over a common set of 6 topic categories [1].

---
[1] http://vikas.sindhwani.org/manifoldregularization.html
[2] http://archive.ics.uci.edu/ml/datasets/Multiple+Features
[3] http://archive.ics.uci.edu/ml/machine-learning-databases/00259/

Table II: Summary of the datasets

| Dataset | # of instances | views | # of clusters |
|---|---|---|---|
| WebKB | 1,051 | Content(3,000), Anchor text (1,840) | 2 |
| Digit | 2,000 | Fourier (76), Profile (216), Karhunen-Loeve (64), Pixel (240), Zernike (47) | 10 |
| Reuters | 111,740 | English (21,531), French (24,893), German (34,279), Spanish (15,506), Italian (11,547) | 6 |
| YouTube | 92,457 | Vision (512), Audio (2,000), Text (1,000) | 31 |

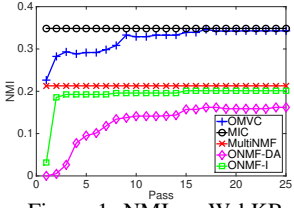
Figure 1: NMI on WebKB
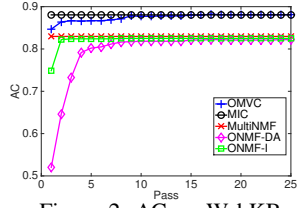
Figure 2: AC on WebKB
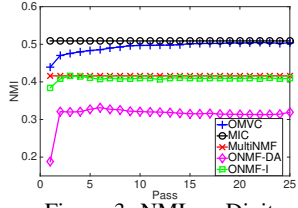
Figure 3: NMI on Digit
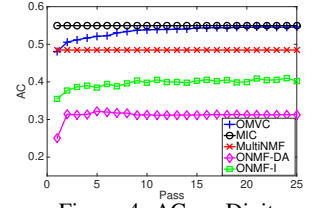
Figure 4: AC on Digit
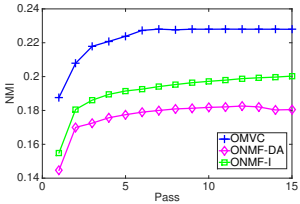
Figure 5: NMI on Reuters
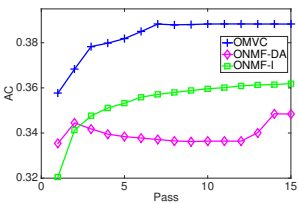
Figure 6: AC on Reuters
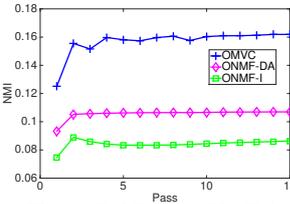
Figure 7: NMI on YouTube
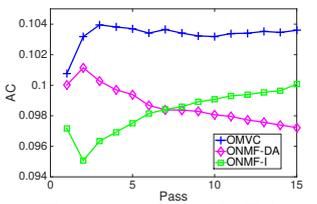
Figure 8: AC on YouTube

Table III: Summary of the comparison methods

| Methods | Multi-view | Incomplete | Sparsity | Online |
|---|---|---|---|---|
| OMVC | ✓ | ✓ | ✓ | ✓ |
| MultiNMF | ✓ | × | × | × |
| MIC | ✓ | ✓ | ✓ | × |
| ONMF-I | × | × | × | ✓ |
| ONMF-DA | × | × | × | ✓ |

- **YouTube Multiview Video Games (YouTube)**[4]: This data contains about 120,000 videos from 31 classes corresponding to 30 popular video games and other games [22]. We select one vision view (512 cuboids histogram features), one audio view (2,000 MFCC features) and one text view (1,000 LDA features from descriptions, titles and tags) in the experiments. We removed some instances with label 31 and kept 92,457 instances to create a more balanced dataset.

### B. Comparison Methods

We compare OMVC with several state-of-the-art methods. The differences between these comparison methods are summarized in Table III, and the details of comparison methods are as follows:

- **OMVC**: OMVC is the proposed online multi-view clustering method in this paper. To facilitate comparison, we set $\alpha_v$s ($\beta_v$s) to be the same for all the views.
- **MultiNMF**: MultiNMF is the state-of-the-art off-line multi-view clustering method based on joint nonnegative matrix factorization [20] for complete views.
- **MIC**: MIC is one of the most recent works that solve the off-line multi-view clustering problem with incomplete data via weighted joint NMF [25].
- **ONMF**: ONMF is an online document clustering algorithm for single view using NMF [30]. In order to apply ONMF, we simply concatenate all the normalized views together to form a big single view. We compare two versions of ONMF from the original paper. **ONMF-I** is the original algorithm that calculates the exact inverse of Hessian matrix, while **ONMF-DA** uses diagonal approximation for the inverse of Hessian matrix.

It is worth mentioning that MIC and MultiNMF are off-line methods which take all data into consideration and can often achieve better performance than on-line methods.

### C. Experiment Setup

In our experiments, two widely used evaluation metrics, accuracy (AC) and normalized mutual information (NMI), are used to measure the clustering performance [31]. Note that all the four datasets are complete. In order to simulate situations with missing instances, we randomly delete instances (0% to 40%) from each view to make the views incomplete. Since all the methods except ONMF have several parameters, we do a grid search for all the parameters in the comparison methods and present the best results obtained. Furthermore, MultiNMF, ONMF-I and ONMF-DA cannot handle incomplete views. In order to apply these methods, we fill the missing instances with average features. In the evaluation, we use $K$-means to get the clustering solution from the consensus latent feature matrix. Since $K$-means depends on initialization, we repeat clustering 20 times with random initialization and report the average performance.

---
[4]https://archive.ics.uci.edu/ml/datasets/YouTube+Multiview+Video+Games+Dataset

Table IV: Run time for different methods

|          | Run Time (seconds) | | | |
|----------|-------|-------|---------|---------|
|          | WebKB | Digit | Reuters | YouTube |
| OMVC/Pass | 16.69 | 27.47 | 1963.82 | 1547.78 |
| ONMF-DA/Pass | 14.53 | 31.08 | 1018.97 | 1201.90 |
| ONMF-I/Pass | 14.39 | 27.26 | 1213.50 | 2589.94 |
| MIC | 1133.23 | 2187.88 | - | - |
| MultiNMF | 974.12 | 747.64 | - | - |

### D. Results

To show the performance of proposed mult-pass OMVC, we randomly deleted 40% of the instances in each view for all the datasets, and run the comparison methods. The chunk size $s$ for online methods is set to 50 for small datasets and 2000 for large datasets. We report both NMI and AC for different passes. The results are shown in Figs. 1-8 and the run times are reported in Table IV. It is worth noting that both MIC and MultiNMF are off-line methods with one pass, so the NMI and AC are two horizontal lines in the figures.

From Figs. 1-4 on WebKB and Digit, we can observe that for the three online methods (OMVC, ONMF-DA and ONMF-I), both NMI and AC increase as the number of passes increases. The performance of the online methods converges as the number of passes increases. Although the off-line method MIC achieves the best performance, the proposed OMVC gets close performance within a few passes and outperforms the other three comparison methods by a large margin. Even in the first pass, the proposed OMVC already outperforms the other two online methods and even the off-line MultiNMF. This shows that even one-pass OMVC can achieve reasonable performance. From these figures, we can see that the comparison methods can be grouped into two groups by the performance. The first group, MIC and OMVC, achieves better performance than the other three methods. It is because that both MIC and OMVC utilize a weight matrix for each view to eliminate the influence of the incomplete data and enforce the sparsity of the latent features, while the other three methods do not consider the incompleteness and sparsity of the data.

Figs. 5-8 demonstrate the performance on the two large data datasets, Reuters and YouTube. However, as the data is too large, the two off-line methods (MIC and MultiNMF) cannot be applied. We only report the NMI and AC for the three online methods. In the four figures, OMVC outperforms the other two online methods in all the passes.

From Figs. 1-8, we can conclude that the proposed OMVC outperforms all the other online methods and perform on par with the best off-line method within a small number of passes. Another interesting observation we can get from the figures is that ONMF-I performs better than ONMF-DA on all the datasets except for YouTube, which indicates that the diagonal approximation sacrifices the accuracy for the computation efficiency in the three datasets.

We also reported the run time for the comparison methods on the four datasets in Table IV. From the table, we can see that all the three online methods are much faster than the two off-line methods. Although the ONMF-DA and ONMF-I run faster than the proposed OMVC, OMVC achieves much better performance than ONMF-DA and ONMF-I.

In OMVC, we need to set the size of data chunk. In order to study the performance of OMVC with different chunk sizes, we conducted another set of experiments. Moreover, to show how the instances incomplete rate affects the performance, We ran the comparison methods with different chunk sizes on WebKB and Digit with different incomplete rates and report the NMI after 10 passes in Table V and Table VI. From the tables, we can first observe that OMVC outperforms the other online methods in all the cases and is very close to the best off-line method, if not better. If we look at the performance for different incomplete rates, we can see that as the incomplete rate increases, the performance for all the methods decreases. It is because as the incomplete rate increases, the useful information contains in each view decreases, and all the methods suffer from the incompleteness of views. We can also observe that for each incomplete rate, when the chunk size is too small (e.g., $s = 2$), all the online methods show low performance. This is because the more data in one chunk, the more information we can use to improve the performance. When $s$ is large enough (larger than $K$), the performance of online methods will improve. From the tables, we can see that when $s$ is 50 or 250, the performance is already pretty close to the best off-line method. Also, using larger chunks means fewer iterations, which reduces the IO cost significantly comparing with using smaller chunks.

### E. Convergence Study

We use the average loss to measure the performance of OMVC after reading each data chunk $t$ in each passes. The average loss is defined as follows:

$$\mathcal{L}^{(t)} = \frac{1}{\min\{s \times t, N\}} \sum_{v=1}^{n_v} \sum_{i=1}^{t} \left( P_i^{(v)} + \alpha_v Q_i^{(v)} + \beta_v \|\mathbf{V}_i^{(v)}\|_1 \right) \tag{28}$$

where $P_i^{(v)} = \|(\mathbf{X}_i^{(v)} - \mathbf{U}^{(v)} \mathbf{V}_i^{(v)^T}) \mathbf{W}_i^{(v)}\|_F^2$ is the reconstruction error for the $i$-th data chunk in $v$-th view and $Q_i^{(v)} = \|\mathbf{W}_i^{(v)}(\mathbf{V}_i^{(v)} - \mathbf{V}_i^*)\|_F^2$ is the distance between the latent features for the $i$-th data chunk in the $v$-th view and the common consensus.

We run OMVC on all the four datasets with 40% incomplete views and report the average loss in Figs. 9-12. From Fig. 9 for WebKB and Fig. 10 for YouTube, we can observe that for each pass, the average loss goes up first and then slowly drops to a certain value. If we compare the lines for different passes, we can see that, at the end of each pass, all the lines converges to one value and the average loss for later pass is lower than the previous pass. From Fig. 11 for Digit and Fig. 12 for Reuters, we can clearly observe that for each pass, as more chunks of data come, the average loss drops and converges. At the end of each pass, the average

Table V: NMI on WebKB with different incomplete rates and chunk sizes

| | 0% | | | | 20% | | | | 40% | | | |
|---|---|---|---|---|---|---|---|---|---|---|---|---|
| | $s=2$ | $s=10$ | $s=50$ | $s=250$ | $s=2$ | $s=10$ | $s=50$ | $s=250$ | $s=2$ | $s=10$ | $s=50$ | $s=250$ |
| OMVC | 0.5852 | 0.5959 | 0.5954 | 0.5529 | 0.2686 | 0.2990 | 0.4339 | 0.4139 | 0.1438 | 0.1966 | 0.3462 | 0.3486 |
| ONMF-DA | 0.1857 | 0.3172 | 0.3689 | 0.3769 | 0.1498 | 0.2407 | 0.2463 | 0.2518 | 0.1322 | 0.1290 | 0.1560 | 0.1903 |
| ONMF-I | 0.4014 | 0.4172 | 0.3428 | 0.2505 | 0.1623 | 0.2337 | 0.2507 | 0.2597 | 0.1290 | 0.133 | 0.188 | 0.2114 |
| MIC | 0.6010 | | | | 0.4013 | | | | 0.3512 | | | |
| MultiNMF | 0.5890 | | | | 0.3884 | | | | 0.2198 | | | |

Table VI: NMI on Digit with different incomplete rates and chunk sizes

| | 0% | | | | 20% | | | | 40% | | | |
|---|---|---|---|---|---|---|---|---|---|---|---|---|
| | $s=2$ | $s=10$ | $s=50$ | $s=250$ | $s=2$ | $s=10$ | $s=50$ | $s=250$ | $s=2$ | $s=10$ | $s=50$ | $s=250$ |
| OMVC | 0.4631 | 0.7203 | 0.7303 | 0.7258 | 0.3240 | 0.6590 | 0.6614 | 0.6527 | 0.2594 | 0.4635 | 0.4885 | 0.4976 |
| ONMF-DA | 0.2762 | 0.2994 | 0.4218 | 0.4195 | 0.3752 | 0.2933 | 0.6118 | 0.3600 | 0.2205 | 0.1473 | 0.3103 | 0.3136 |
| ONMF-I | 0.2545 | 0.3142 | 0.6735 | 0.6688 | 0.1892 | 0.1792 | 0.3318 | 0.3102 | 0.2588 | 0.3146 | 0.3785 | 0.4138 |
| MIC | 0.7305 | | | | 0.6569 | | | | 0.4903 | | | |
| MultiNMF | 0.7313 | | | | 0.6412 | | | | 0.4304 | | | |

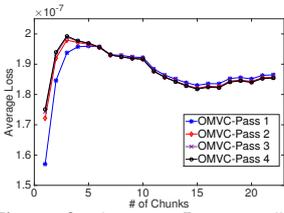
Figure 9: Average Loss vs # of chunks on WebKB

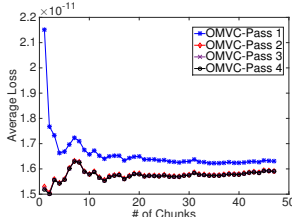
Figure 10: Average Loss vs # of chunks on YouTube

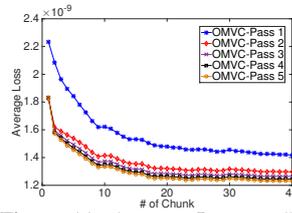
Figure 11: Average Loss vs # of chunks on Digit

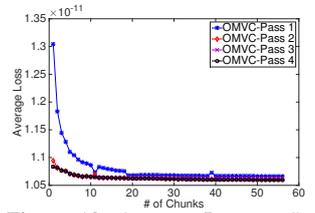
Figure 12: Average Loss vs # of chunks on Reuters

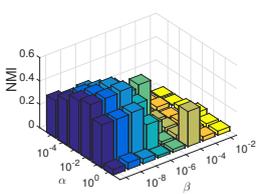
(a) Digit

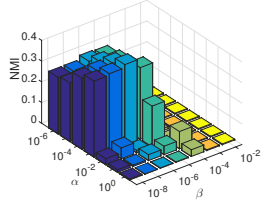
(b) WebKB

Figure 13: Sensitivity analysis.

losses for later passes are lower than previous passes and converge to a certain value.

*F. Sensitivity Analysis*

There are two sets of parameters in the proposed methods: $\{\alpha_v\}$ and $\{\beta_v\}$. Here we explore the effects of the two parameter sets. As mentioned in the previous section, we set $\alpha_v$ ($\beta_v$) to be the same for all the views. For simplicity, assume that $\alpha_v = \alpha$ and $\beta_v = \beta$ for all the views. We run OMVC with different values for $\alpha$ and $\beta$ on WebKB and Digit data. To save space, we only show the results w.r.t. NMI in Fig. 13 since we have similar observations in AC.

From Fig. 13a, we can observe that OMVC achieves the best performance when $\alpha$ is about $10^{-2}$ and $\beta$ is about $10^{-7}$. Parameter $\alpha$ controls the importance of co-regularization between views and the consensus. When it becomes too small, the consensus has little contribute to the learning of each view, and the performance will decline. When $\alpha$ increases, the consensus has too much influence to each view and the performance will drop. Parameter $\beta$ controls the sparsity of the latent feature matrices. We can observe that when it is too small, we barely enforce the sparsity, and thus the performance decline. When it is too large, most of the entries in the latent feature matrices will be 0, and the performance will drop. We can have similar observation from Fig. 13b. These results show that an appropriate combination of these two parameters is crucial for OMVC to improve the performance.

## VI. RELATED WORK

Multi-view clustering [4] provides a natural way for generating clusters from multi-view data. In the introduction, we have discussed four categories of multi-view clustering algorithms. Here we particularly address several subspace based multi-view clustering methods [8, 20, 25, 29]. [8] proposed a CCA based multi-view clustering method to learn the subspace, in which the correlations among views are maximized. [20] approached the problem by learning a common consensus based on a co-regularized joint NMF framework. Recently, [29] proposed to solve multi-view clustering with at least one complete view based on CCA. Later, [25, 17] proposed two NMF based methods that can solve multi-view clustering even without any complete view. Although various methods have been proposed to integrate heterogeneous views, none of the previous methods can handle large-scale data that cannot fit into the memory.

Nonnegative matrix factorization [16], especially online NMF is the second area that is related to our work. Since traditional NMF cannot deal with really large data, online

NMF was first proposed to handle really large data or streaming data [6]. Different variations were proposed in the last few years. For example, [11] proposed an efficient online NMF algorithm that takes one chunk of samples per step and updates the bases via robust stochastic approximation. [30] proposed an online NMF algorithm for document clustering. It utilizes the second-order Hessian matrix to optimize the objective incrementally. [24] added graph regularization to an online joint NMF framwork for multi-view feature selection. However, none of them can deal with multiple incomplete views. OMVC uses a weighted joint NMF model to handle the incompleteness of the views and enforces the sparseness of the learned latent feature matrices.

## VII. CONCLUSION

In this paper, we present possibly the first attempt to solve the online multi-view clustering with incomplete views where each view may suffer from missing some instances. Based on NMF, the proposed OMVC learns the latent feature matrices for each individual incomplete view and pushes them towards a common consensus. To achieve the goal, a joint NMF algorihm is used to not only incorporate individual matrix factorizations but also minimize the disagreement between the latent feature matrices and the consensus. By giving missing instances lower weights dynamically, OMVC minimizes the negative influences of the missing data. OMVC also enforces the sparsity of the learned latent feature matrices by introducing lasso regularization, which makes the method robust to noises and outliers. Most important, OMVC does not require holding the entire data matrix into memory, which reduces the space complexity dramatically. It processes the data one by one (or chunk by chunk), learns the latent feature and updates the basis matrix simultaneously. Extensive experiments conducted on both small and large real data demonstrate the effectiveness of the proposed OMVC method comparing with other state-of-the-art methods.

## ACKNOWLEDGMENT

This work is supported in part by NSF through grants IIS-1526499, CNS-1626432, NSFC through grants 61503253, 61672357, and NVIDIA Corporation with the donation of the Titan X GPU used for this research.